\documentclass[conference]{IEEEtran}
\IEEEoverridecommandlockouts
\usepackage{cite}
\usepackage{amsmath,amssymb,amsfonts}
\usepackage{algorithmic}
\usepackage{graphicx}
\usepackage{textcomp}
\usepackage{xcolor}
\usepackage{soul}
\usepackage{multirow}
\def\BibTeX{{\rm B\kern-.05em{\sc i\kern-.025em b}\kern-.08em
		T\kern-.1667em\lower.7ex\hbox{E}\kern-.125emX}}

\begin{document}
	\title{{A Slow Shifting Concerned Machine Learning Method for Short-term Traffic Flow Forecasting}\\
		\thanks{This research is supported by A*STAR under its RIE2020 Advanced Manufacturing and Engineering (AME) Industry Alignment Fund – Pre Positioning (IAF-PP) (Grant No. A19D6a0053). Any opinions, findings and conclusions, or recommendations expressed in this material are those of the author(s) and do not reflect the views of A*STAR.}
	}
    \makeatletter
	\newcommand{\linebreakand}{%
          \end{@IEEEauthorhalign}
          \hfill\mbox{}\par
          \mbox{}\hfill\begin{@IEEEauthorhalign}
        }
        \makeatother
	\author{\IEEEauthorblockN{Zann Koh}
		\IEEEauthorblockA{\textit{Engineering and Product Development} \\
			\textit{Singapore University of Technology and Design}\\
			Singapore, Singapore \\
			zann\_koh@mymail.sutd.edu.sg}
		\and
		\IEEEauthorblockN{Yan Qin}
		\IEEEauthorblockA{\textit{Engineering and Product Development} \\
			\textit{Singapore University of Technology and Design}\\
			Singapore, Singapore \\
			zdqinyan@gmail.com}
		\linebreakand
		\IEEEauthorblockN{Yong Liang Guan}
		\IEEEauthorblockA{\textit{School of Electrical and Electronics Engineering} \\
			\textit{Nanyang Technological University}\\
			Singapore, Singapore \\
			eylguan@ntu.edu.sg}
		\and
		\IEEEauthorblockN{Chau Yuen}
		\IEEEauthorblockA{\textit{Engineering and Product Development} \\
			\textit{Singapore University of Technology and Design}\\
			Singapore, Singapore \\
			yuenchau@sutd.edu.sg}
	}
	
	\maketitle
	
	\begin{abstract}
		The ability to predict traffic flow over time for crowded areas during rush hours is increasingly important as it can help authorities make informed decisions for congestion mitigation or scheduling of infrastructure development in an area. However, a crucial challenge in traffic flow forecasting is the slow shifting in temporal peaks between daily and weekly cycles, resulting in the nonstationarity of the traffic flow signal and leading to difficulty in accurate forecasting. To address this challenge, we propose a slow shifting concerned machine learning method for traffic flow forecasting, which includes two parts. First, we take advantage of Empirical Mode Decomposition as the feature engineering to alleviate the nonstationarity of traffic flow data, yielding a series of stationary components. Second, due to the superiority of Long-Short-Term-Memory networks in capturing temporal features, an advanced traffic flow forecasting model is developed by taking the stationary components as inputs. Finally, we apply this method on a benchmark of real-world data and provide a comparison with other existing methods. Our proposed method outperforms the state-of-art results by 14.55\% and 62.56\% using the metrics of root mean squared error and mean absolute percentage error, respectively.
	\end{abstract}
	
	\begin{IEEEkeywords}
		Traffic flow forecasting, Long-short term memory, Empirical mode decomposition
	\end{IEEEkeywords}
	
	\section{Introduction}
	With the recent improvement of technologies in sensing as well as data analysis, there are an increasing number of real-world applications, one of which is the field of smart mobility. Most humans need to travel from place to place daily, whether for work commute or leisure. To accommodate and ensure smoothness of travel, the local governments would need to have a clearer picture of the travel patterns of the general population. If the traffic flow at a certain location over different times of day can be predicted, then authorities can take measures against events such as congestion or traffic jams, or even schedule road works or expansions to adapt to the traffic flow in the relevant regions.
	
	With a large amount of data currently available to be gathered for traffic flow, it is feasible to examine the use of advanced machine learning algorithms that can be used to make sense of such data. Various machine learning algorithms have been reported in traffic flow forecasting. However, the most common choice of recurrent neural network (RNN) used in the field of traffic flow forecasting is the Long-Short Term Memory (LSTM) network according to Tedjopurnomo \textit{et al.} \cite{tedjopurnomo2020survey}. Literature using LSTM networks makes up 18 out of the 37 surveyed works using RNN in their paper. LSTM has also been extensively used in other fields such as indoor temperature modeling \cite{elmaz2021cnn}, charge estimation for battery health \cite{qin2021transfer}, forecasting tourism demand \cite{he2021using}, as well as for prediction of the next location of a vehicle's trajectory \cite{qin2023spatiotemporal}. An additional advantage of LSTM is that it is easily transferable \cite{qin2022transferable}.
    
	Works using LSTM in traffic flow forecasting include the work by Zou \textit{et al.} \cite{zou2018city} where they adopted a simple LSTM model to predict the citywide traffic flows. Luo \textit{et al.} \cite{luo2019spatiotemporal} use a combination of $k$-nearest neighbors (KNN) and LSTM in their work to predict the traffic flow of roads at certain detector stations located along the roads of interest. They applied KNN to select a fixed number of nearest stations that are closely related to the target station. Using the data of those related stations in their model, the prediction accuracy is improved. Shin \textit{et al.} \cite{shin2020prediction} and Tian \textit{et al.} \cite{tian2018lstm} made use of the patterns found in missing data as part of their application of LSTM in traffic prediction. LSTM has also been used in conjunction with other neural networks for improved forecasting performance, such as with graph neural networks by Lu \textit{et al.} \cite{lu2020lstm} and with convolutional neural networks by Liu \textit{et al.} \cite{liu2017short} and Wang \textit{et al.} \cite{wang2021effective}.
	
	In \cite{yao2019revisiting}, LSTM and a spatial convolutional neural network were used to harness the additional information in the spatial dimension and improve the prediction performance. They also made use of attention mechanisms to capture temporal shifting in peaks from day to day and week to week. Attention mechanisms have also been used in works such as those by Zhao \textit{et al.} \cite{zhao2020attention}, Yang \textit{et al.} \cite{yang2019traffic}, and Guo \textit{et al.} \cite{guo2021ma} in conjunction with LSTMs for improvement in prediction results. Although the traffic forecasting algorithms proposed in the aforementioned works perform well, they mostly include other external data sources to aid their prediction, such as spatial data and weather data. However, this kind of external information may be difficult to obtain. Our target is to come up with a traffic flow forecasting method that does not rely on external data, yet has comparable results to such methods.
 
	To improve the performance of classical LSTM, feature engineering can also be applied. There are various forms of feature engineering such as calculating a meaningful index \cite{yi2019implementing}, using canonical variate analysis \cite{zhou2022transfer}, or decomposition of the raw signal \cite{liu2020short}. To determine the most appropriate feature engineering method, we examine the properties of the relevant data. Daily traffic flow data roughly approximates a periodic nature but is not exactly periodic, due to the slow shifting of temporal peaks in daily and weekly cycles. The challenge of this slow shifting nature is present in other fields as well \cite{qin2022slow}. To address this slow shifting nature, we decided to explore the usage of Empirical Mode Decomposition (EMD). EMD is a method for finding an intuitive representation of the different frequency components within complicated and dynamic signals. In the case of representing the daily variations in traffic flow, EMD is more advantageous over the commonly used Fourier transform \cite{bracewell1986fourier} as traffic flow is not perfectly periodic. There can be general daily patterns such as `peak in the morning, valley in the evening' but these peaks and valleys do not necessarily occur at the same timing of every day or even the same day of every week. Representing these dynamic signals with the Fourier transform may require a large number of different frequencies of sinusoidal basis functions. With EMD, the fluctuations within a time series are automatically and adaptively selected from the time series itself, without requiring a predetermined set of mathematical functions.
	
	EMD has also been used to improve results in several time series prediction applications. Tian \textit{et al.} used it in their work \cite{tian2021emd} to extract Intrinsic Mode Functions (IMFs) from simulated network traffic flows and combine them with the Autoregressive Moving Average (ARMA) algorithm for prediction in the field of Internet of Vehicles. In the field of hydrology, it has been used by Agana and Homaifar \cite{agana2018emd} with deep belief networks for drought forecasting. For traffic flow forecasting, Feng \textit{et al.} \cite{feng2015prediction} used it in conjunction with wavelet neural networks, while Chen \textit{et al.} \cite{chen2021traffic} used it as part of an ensemble framework that includes deep learning. However, in these cases, they have mainly utilized EMD as a means of noise removal rather than as a feature extraction method. In the case of the work by Wang \textit{et al.} \cite{wang2016novel}, EMD was used to decompose the original time series signal and used multiple AutoRegressive Integrated Moving Average (ARIMA) to forecast each decomposed IMF individually. Other works including those by Hao et al \cite{hao2022hybrid} and Chen et al \cite{chen2019hybrid} use LSTM to separately forecast the decomposed IMFs, then combined the forecasting results of all the IMFs to form the final result. 
	
	In the present research, there are many works making use of LSTM and EMD for time series prediction. However, to the best of the authors' knowledge, it is a pointer to utilize EMD as a feature extraction method instead of a denoising method and combine it with an LSTM model for the prediction of traffic flow. In this study, we propose a method named Slow-shifting Temporal Traffic flow Forecasting (STTF). STTF makes use of the IMFs obtained through the process of EMD to extract features from a given traffic flow data at a given location for machine learning. These features are then fed into an LSTM model and the output of the model is the forecast value of the traffic volume at the next time interval, which is different from those in previous works which use LSTM to predict the next values of the IMFs and combine those predicted values, rather than directly predict the traffic flow value at the next interval, which is what we have done. Our method aims to improve the prediction results for the traffic flow of the next time interval in terms of root-mean-squared error (RMSE) as well as mean absolute percentage error (MAPE). We show that our method gives comparable results to state-of-the-art technology while making use of only the temporal data, thus reducing the need for external sources of data.
	
	The remainder of the paper is organized as follows: Section II describes the adopted dataset. Section III details our proposed methodology, the Slow-shifting Temporal Traffic flow Forecasting. The experimental results and comparison of Slow-Shift Temporal Prediction to other methods are shown in Section IV. Finally, Section V concludes the paper.
    
    \section{Dataset Description}
	To perform traffic flow forecasting, having a good source of mobility data is crucial. One viable source of mobility data for the prediction of traffic flow is taxis that are equipped with Global Positioning System (GPS) receivers. Such taxicabs generate large volumes of data daily, as noted in the work of Zheng \textit{et al.} \cite{zheng2011urban}, in which they state that there are around 67,000 licensed taxicabs that generate over 1.2 million occupied trips per day in Beijing alone. If actionable mobility insights can be extracted from these large volumes of taxi data, it would save governing bodies time and resources which they would otherwise have to spend on installing specific infrastructure and collecting specific data.
    
    For this study, we used the NYC taxi dataset, which was collected by the New York Taxi Limousine Company\footnote{https://www.nyc.gov/site/tlc/about/tlc-trip-record-data.page} and processed by Yao \textit{et al.} \cite{yao2019revisiting}. This dataset represents the end flow volume at each 1km-by-1km square in their 10km-by-20km selected study area in 30min intervals over 60 days. The first 40 days of data were used as the training set, while the remaining 20 days were used as the testing set.
    
    \begin{table}[htbp]
    	\caption{Details of dataset used in this study}
    	\centering
    	\def\arraystretch{1.5}
    	\begin{center}
    		\begin{tabular}{p{0.4\linewidth}|c }
    			\hline
    			\hline
    			\textbf{Dataset Characteristics} & \textbf{Value} \\
    			\hline
    			\textbf{Time span (DD/MM/YYYY)} & 01/01/2015 - 01/03/2015 \\
    			\hline
    			\textbf{Time interval} & 30 min \\
    			\hline
    			\textbf{Range of traffic flow values per interval (min – max)}  & \multirow{2}{*}{0 - 954} \\        
    			\hline
    			\textbf{Median traffic flow value} & 431 \\        
    			\hline
    			\hline
    		\end{tabular}
    		\label{tab:data}
    	\end{center}
    \end{table}
	
	\begin{figure*}[tbp]
		\centerline{\includegraphics[width=0.8\linewidth]{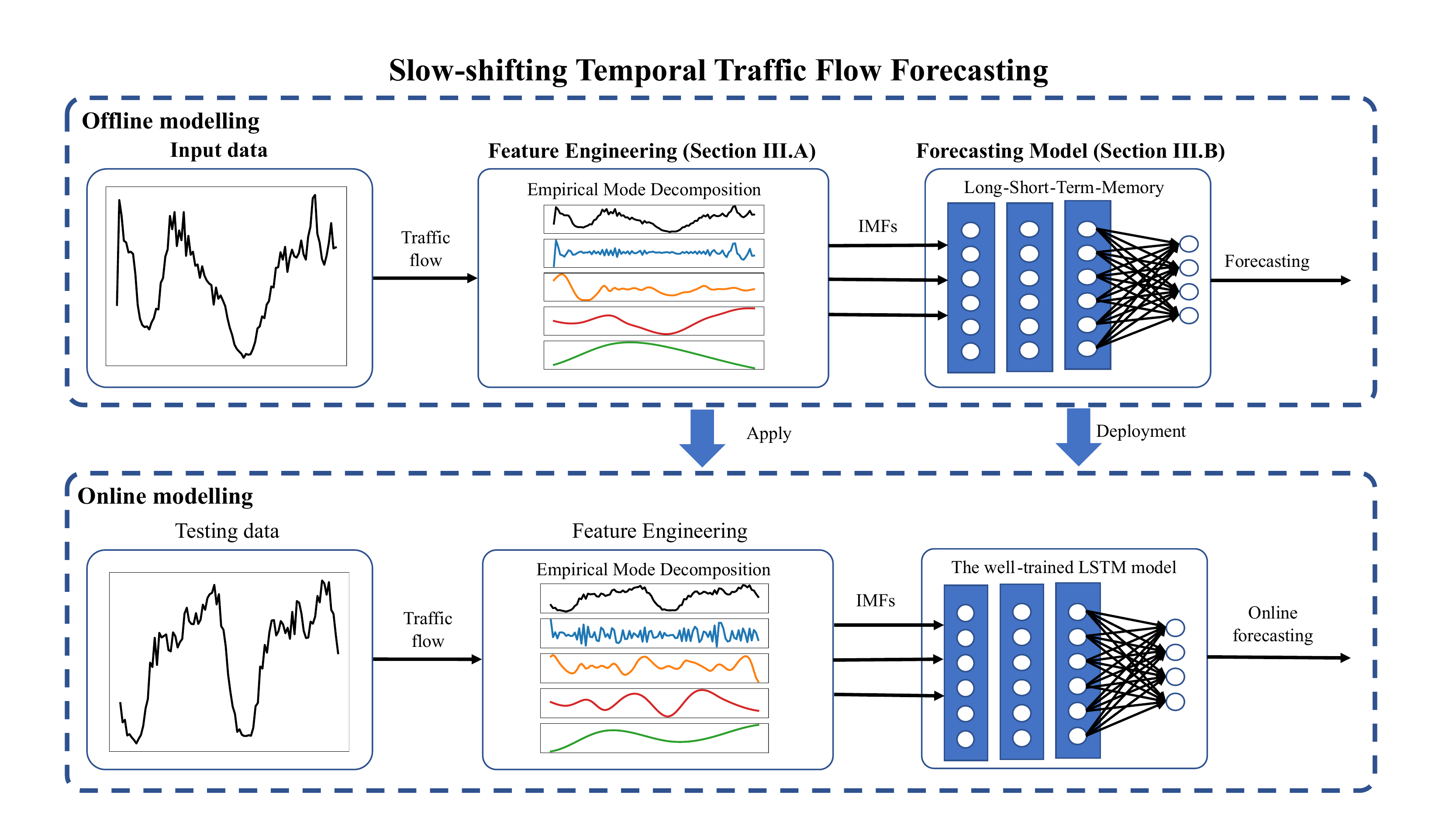}}
		\caption{The overall structure of the proposed Slow-shifting Temporal Traffic flow Forecasting.}
		\label{fig:flowchart}
	\end{figure*}
	
	For our study, the time series data of the single 1km-by-1km square with the highest average traffic flow value was selected. To help the signal be more compatible with EMD, we took the difference between the mean of the signal and the original signal and used that as an input for EMD. The mean value of the signal was noted and added back to the predicted value afterward, before comparing it with the original labels of the testing dataset.
	
	To prepare the data for the LSTM input, we used a sliding window to extract the values of two time steps before as well as the current time step as the features and the value of the next time step as the label. This means that for a time $t$, we have the values of all the separated IMFs at times $t-2$, $t-1$, and $t$ as the features, and we want to predict the actual value of the traffic flow at time $t+1$.

	\section{Slow-shifting Temporal Traffic Flow Forecasting}
	This section introduces our proposed methodology as well as a brief overview of the algorithms used. The overall structure of our data flow is illustrated in Fig.~\ref{fig:flowchart}.  To address the slow shifting nature of the traffic flow data, the time series data is firstly decomposed into its constituent Intrinsic Mode Functions (IMFs) using the process of EMD. The separated IMFs are then used as individual features and fed into an LSTM model with attention layers. For each sample, the features comprise the values of each IMF at that timestep, as well as the IMFs at the timestep before that, up to 2 prior timesteps in this study. Thus, we call our proposed method Slow-shifting Temporal Traffic flow Forecasting (STTF).

	\subsection{Feature Engineering}
        One of the main challenges in traffic flow forecasting is the slow shifting in the peak locations in the time signal of the traffic flow data. Traffic may peak at slightly different timings each day, and this slow shifting in peak timings leads to nonstationarity, which contributes to the difficulty of prediction. To address this, we choose to use EMD. EMD was proposed by Huang \textit{et al.} \cite{huang1998empirical} in order to separate the representations of different frequency components within a signal or time series.
	
	EMD operates under the following assumptions: 
	\begin{itemize}
		\item The data has at least two extreme values, one maximum, and one minimum.
		\item The local time-domain characteristics of the raw data are uniquely determined by the time scale between the extreme points.
		\item If there is no extreme point in the data but there is an inflection point, the result can be obtained by taking the derivative of the data one or more times to obtain the extreme value and then integrating it.
	\end{itemize}

	The different frequency components within a signal or time series are called Intrinsic Mode Functions (IMFs). Each IMF has to satisfy the following conditions:
	\begin{itemize}
		\item The total number of local extrema (local maxima and local minima) and the number of zero-crossings must either be equal or differ at most by one.
		\item At any point, the mean value of the envelope defined by the local maxima and the envelope defined by the local minima is near zero.
	\end{itemize}
	
	The EMD process of extracting IMFs from a time signal $X(t)$ is as follows. Firstly, all the local extrema (maxima and minima) are located. Next, all the local maxima are connected by a cubic spline line to form an upper envelope $e_{up}(t)$. A similar process happens with all the local minima to obtain a lower envelope $e_{lo}(t)$. These two envelopes should cover all the data in between. 
	
	Next, a mean value $m(t)$ is computed by taking the mean of these two envelopes. 
	
	\begin{equation}
		m(t) = (e_{up}(t) + e_{lo}(t)) / 2
	\end{equation}
	
	The first test component $d(t)$ is then extracted as the difference between the original data $X(t)$ and the mean of the envelopes $m(t)$. 
	\begin{equation}
		d(t) = X(t) - m(t)
	\end{equation}
	
	At this point, the properties of $d(t)$ are checked to determine if $d(t)$ fulfills the conditions of an IMF. If it does, $d(t)$ is taken to be an IMF, and the original signal $X(t)$ is replaced by the residual $r(t) = X(t) - d(t)$ for further computation of subsequent IMFs. If it does not, then $X(t)$ is replaced directly by $d(t)$ for further computation. This iterative process continues until the last residual function becomes a monotonic function or the number of extrema is less than or equal to one. This means that no more IMFs can be extracted.
	
	A plot of IMFs that were extracted via EMD is shown in Fig.~\ref{fig:emd}. The signal in black at the top is sum of all the IMFs which ideally would reproduce the original signal. The individual component IMFs are plotted in the rows below from highest to lowest frequency.
	
	\begin{figure}[tbp]
		\centerline{\includegraphics[width=\linewidth]{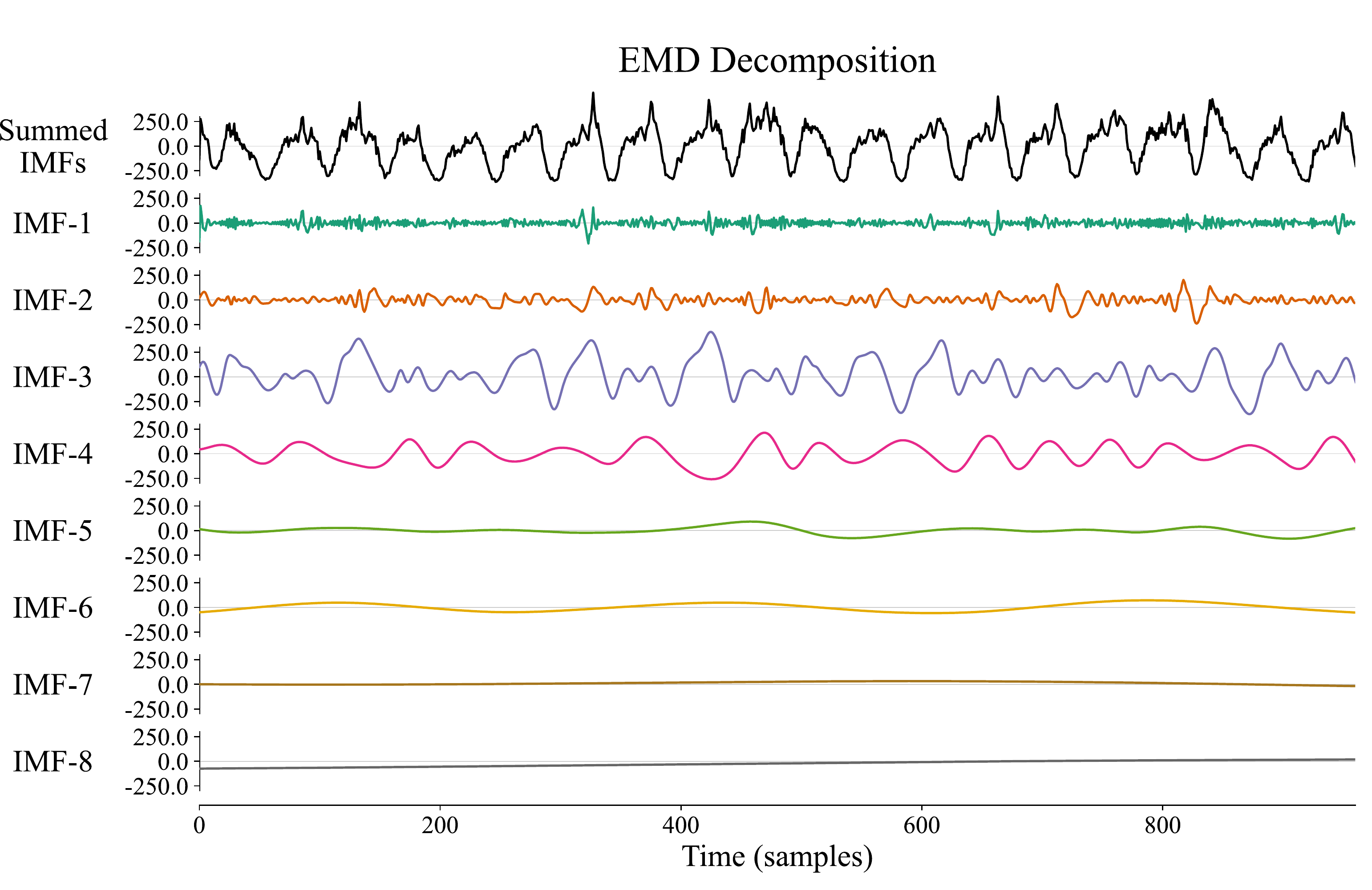}}
		\caption{Decomposition of a sample time signal from the data using EMD. Summed IMFs in the top row in black correspond to the original signal.}
		\label{fig:emd}
	\end{figure}	
 
	\subsection{Development of the Forecasting Model with LSTM}
    After obtaining IMFs as features, the next step is to develop a predictive learning model. LSTM is one such predictive model that is particularly suited to time-series predictions. The LSTM was proposed by Hochreiter and Schmidhuber \cite{hochreiter1997long} to improve the long-term temporal prediction ability from a basic RNN. The LSTM targets the problem of vanishing gradients for long-term predictions in RNNs by allowing the `forgetting' or ignoring of data that is not useful for the prediction in the network.
	
	A diagram of an LSTM unit is shown in Fig.~\ref{fig:lstm}. The LSTM unit has three gates - a forget gate, an input gate, and an output gate, denoted by the $f$, $i$, and $o$, respectively, in the following equations. $W$ and $b$ represent weights and biases, respectively. $\sigma$ represents a sigmoid function and the subscript $t$ denotes the value at time $t$.
	\begin{figure}[tbp]
		\centerline{\includegraphics[width=\linewidth]{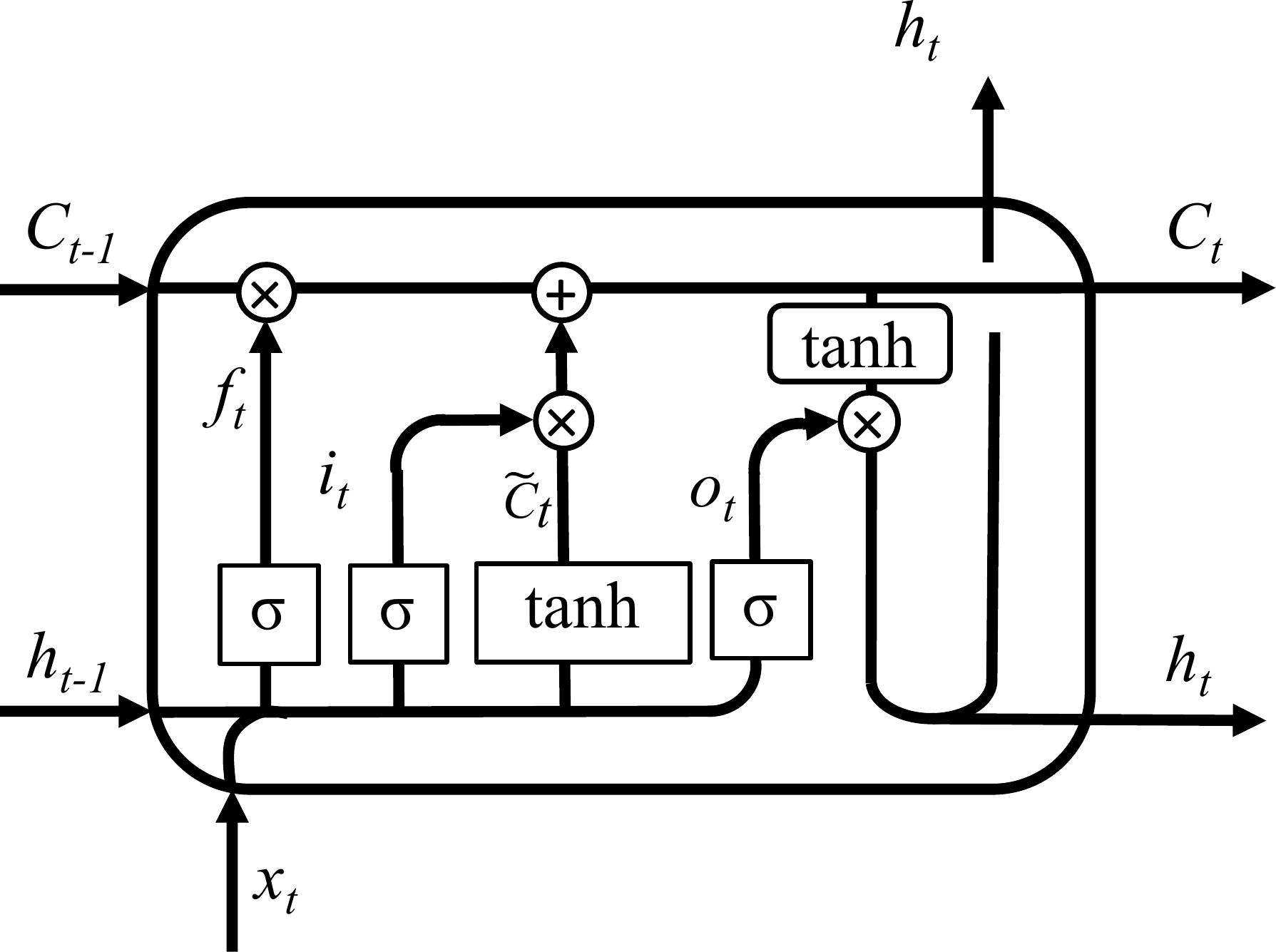}}
		\caption{Diagram of an LSTM unit.}
		\label{fig:lstm}
	\end{figure}	
	
	Firstly, the forget gate decides which parts of the previous cell state to forget by computing $f_t$. 
	\begin{equation}
		f_t = \sigma (W_f \cdot [h_{t-1},x_t] + b_f)
	\end{equation}
	
	The new information from the current sample to store in the cell state is decided by the input state $i_t$.

	\begin{equation}
		i_t = \sigma (W_i \cdot [h_{t-1},x_t] + b_i)
	\end{equation}
	
	Next, a vector of new candidate values for the cell state, $\tilde{C}_t$, is created. 
	
	\begin{equation}
		\tilde{C}_t = \tanh(W_C \cdot [h_{t-1},x_t] + b_C)
	\end{equation}
	
	The cell state $C_t$ is then updated with the summed values of the convolution of $f_t$ with the previous cell state $C_{t-1}$ as well as the convolution of $i_t$ with the candidate cell state vector $\tilde{C}_t$. 
		
	\begin{equation}
		C_t = f_t \ast C_{t-1} + i_t \ast \tilde{C}_t
	\end{equation}
	
	The output gate decides which parts of the information to be outputted. 
	
	\begin{equation}
		o_t = \sigma (W_o \cdot [h_{t-1},x_t] + b_o)
	\end{equation}

	Lastly, the hidden state is updated with the convolution of $o_t$ with the tanh function output of the cell state $C_t$. 
	 
	\begin{equation}
		h_t = o_t \ast \tanh(C_t)
	\end{equation}

	For this study, we used a sequential model with two LSTM layers and an attention layer after each LSTM layer, followed by a densely connected layer and an output layer. The number of neurons in each of the LSTM and dense layers were tested from a set of different numbers of neurons. We found that ten neurons in each LSTM layer, as well as the densely connected layer gave the best result.

	\section{Experimental Results and Discussion}
	This section presents the results of performing STTF on our test dataset as well as the results of comparison against previous time series methods and state-of-the-art methods.
	
	For this experiment, the time series data from the end flow of the grid square with the highest average traffic flow was used. STTF and each of the other methods were used to forecast the values of each subsequent timestep for the length of the testing data. A visual prediction comparison between the proposed STTF and pure LSTM is shown in Fig.~\ref{fig:predictions}. From Fig.~\ref{fig:predictions}, it can be observed that although the pure LSTM can detect the locations of the fluctuations reasonably well, its predictions still have a larger margin of error as compared to STTF.

	\begin{figure}[tbp]
		\centerline{\includegraphics[width=\linewidth]{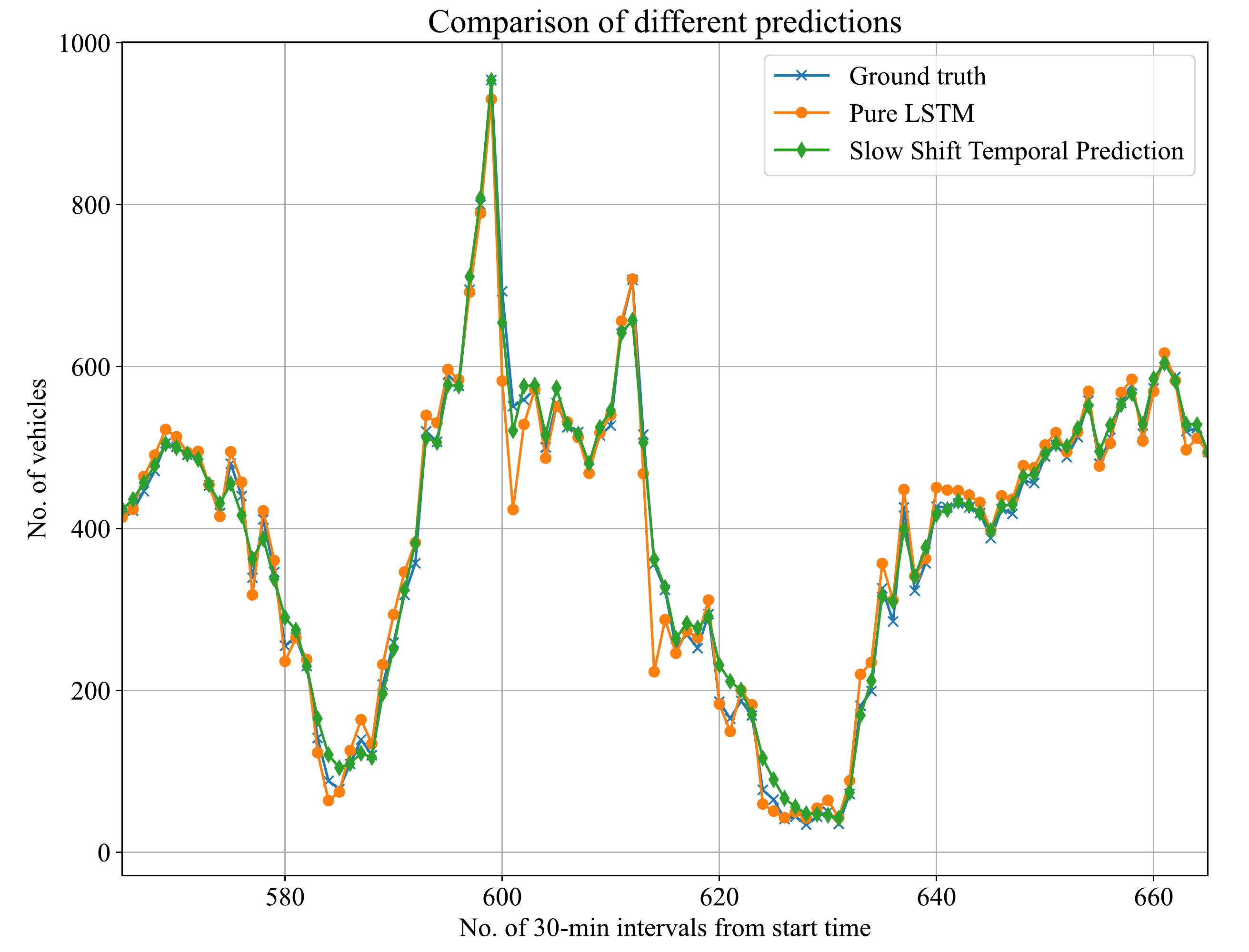}}
		\caption{Forecasting trend comparison between the proposed method and LSTM using the data from time intervals 565 to 665.}
		\label{fig:predictions}
	\end{figure}
	
	For numerical comparison of model performance, the metrics root-mean-squared-error (RMSE) and mean absolute percentage error (MAPE) are selected. The definitions of RMSE and MAPE are shown in  (\ref{eq:rmse}) and (\ref{eq:mape}), respectively. In these equations, $i$ refers to the time step of the prediction, $n$ is the total number of predictions made, $y_{true,i}$ is the true value at the time step $i$, and $y_{pred,i}$ is the predicted value at the time step $i$.
	
	\begin{equation}
		RMSE = \sqrt{\frac{1}{n} \sum_{i=1}^{n}(y_{true,i} - y_{pred,i})^2}
		\label{eq:rmse}
	\end{equation}
	
	\begin{equation}
		MAPE = \frac{1}{n} \sum_{i=1}^{n} \left\lvert \frac{y_{true,i} - y_{pred,i}}{y_{true,i}} \right\rvert \times 100
		\label{eq:mape}
	\end{equation}
	
	For both of these metrics, a lower value indicates better performance as the error is lower. We perform a comparison with the state-of-the-art STDN model, against pure LSTM without the EMD portion, as well as two well-known time series prediction algorithms, ARMA and ARIMA \cite{box2015time}. The respective RMSE and MAPE values for each model type are shown in Table~\ref{tab1}.
	
	\begin{table}[htbp]
		\caption{Comparison between the proposed method and its counterparts regarding prediction accuracy}
		\centering
		\def\arraystretch{1.5}
		\begin{center}
			\begin{tabular}{c|c|c}
				 \hline
                   \hline
				  \textbf{Model} & \textbf{RMSE} & \textbf{MAPE (\%)}\\
				 \hline
				 \textbf{ARMA} & 315.19 & 73.51\\
				 \hline
				 \textbf{ARIMA} & 187.29 & 96.42\\
				 \hline
     		 \textbf{LSTM} & 57.53 & 15.13\\
				 \hline
				 \textbf{STDN \cite{yao2019revisiting}} & 19.05 & 15.60 \\        
        		 \hline
				 \textbf{STTF (Proposed)} & 16.25 & 5.84 \\
				 \hline
				 \hline
			\end{tabular}
			\label{tab1}
		\end{center}
	\end{table}
	
	We can observe that the proposed STTF model outperforms both the STDN as well as the pure LSTM model. The pure LSTM model may not have been able to capture the full extent of the peaks and valleys, thus resulting in a higher RMSE value. The ARMA and ARIMA models performed relatively poorly, which may have been caused by the predictions flatlining due to an extended prediction horizon.
	
	\section{Conclusion}
	In this paper, we have proposed a method for traffic flow forecasting called Slow-shifting Temporal Traffic flow Forecasting that aims to address the challenge of nonstationarity in the traffic flow time signal. We incorporate EMD as a means of feature extraction by extracting the IMFs of the traffic flow data to serve as feature inputs of each sample. LSTM with attention layers is then used as a prediction algorithm. The proposed method was performed on a benchmark dataset and performed better in terms of RMSE and MAPE as compared to the state-of-the-art and some other common algorithms. The proposed method can be applied to time series data from different locations and at different granularities.
	
	\bibliographystyle{IEEEtran}  
	\bibliography{bibliography}

\end{document}